\documentclass[lettersize,journal]{IEEEtran}
\usepackage{amsmath,amsfonts}
\usepackage{algorithmic}
\usepackage{algorithm}
\usepackage{array}
\usepackage{textcomp}
\usepackage{subfigure}
\usepackage{url}
\usepackage{verbatim}
\usepackage{graphicx}
\usepackage{cite}
\usepackage{tabularx}
\usepackage{booktabs}
\usepackage{bm}
\usepackage{orcidlink} 
\begin{document}

\title{Pyramid Token Pruning for High-Resolution Large Vision-Language Models via Region, Token, and Instruction-Guided Importance}

\author{Yuxuan Liang\textsuperscript{$\ast$}$^{\orcidlink{0009-0004-5039-7568}}$, Xu Li\textsuperscript{$\ast$}$^{\orcidlink{0009-0001-2431-3410}}$, Xiaolei Chen$^{\orcidlink{0009-0005-5700-9326}}$, Haotian Chen$^{\orcidlink{0009-0001-0593-5281}}$, Yi Zheng$^{\orcidlink{0009-0006-1549-6979}}$, \\ Zhe Liu$^{\orcidlink{0009-0003-3668-6518}}$, Rui Zhu$^{\orcidlink{0009-0007-7226-6923}}$, Bin Li$^{\orcidlink{0000-0002-9633-0033}}$, Xiangyang Xue\textsuperscript{$\dagger$}$^{\orcidlink{0000-0002-4897-9209}}$, Member, IEEE
\thanks{This work is supported by Shanghai Key Laboratory of Intelligent Information Processing and College of Computer Science and Artificial Intelligence, Fudan University. \textbf{$\ast$ denotes equal contribution. $\dagger$denotes the corresponding author}.}


\thanks{Y Liang, X Li, X. Chen, H. Chen, Y Zheng, C Lai, B Li and X. Xue are with the College of Computer Science and Artificial Intelligence, Fudan University, Shanghai 200433, China (E-mails: {yxliang25, xu\_li23, chenxl23, htchen24, zhengy23, zheliu24, rzhu24}@m.fudan.edu.cn, libin@fudan.edu.cn, xyxue@fudan.edu.cn)}

}

\markboth{Journal of \LaTeX\ Class Files,~Vol.~xx, No.~x, September~202x}%
{How to Use the IEEEtran \LaTeX \ Templates}

\maketitle
\begin{abstract}
Large Vision-Language Models (LVLMs) have recently demonstrated strong multimodal understanding, yet their fine-grained visual perception is often constrained by low input resolutions. A common remedy is to partition high-resolution images into multiple sub-images for separate encoding, but this approach drastically inflates the number of visual tokens and introduces prohibitive inference overhead. To overcome this challenge, we propose Pyramid Token Pruning (PTP), a training-free strategy that hierarchically integrates bottom-up visual saliency at both region and token levels with top-down instruction-guided relevance. Inspired by human visual cognition, PTP selectively preserves more tokens from salient regions while further emphasizing those most relevant to task instructions. Extensive experiments on 13 diverse benchmarks show that PTP substantially reduces computational cost, memory usage, and inference latency, with negligible performance degradation. 
\end{abstract}


\section{Introduction}
\IEEEPARstart{L}{arge} Vision-Language Models (LVLMs) have substantially advanced multimodal understanding by coupling Large Language Models (LLMs) with visual encoders. Early LVLMs such as LLaVA \cite{liu2024llava}, InstructBLIP \cite{instructblip}, and MiniGPT-4 \cite{zhu2024minigpt4} demonstrated impressive image comprehension capabilities, yet they typically operate on low-resolution inputs and thus struggle to capture fine-grained visual details \cite{arif2025hired}. To mitigate this limitation, recent works \cite{liu2024llavanext,chen2024farinternvl1_5} introduced dynamic image partitioning, in which a high-resolution image is divided into multiple sub-images alongside a downsampled global image. This strategy preserves more detailed information—yielding up to a 15$\%$ accuracy gain when scaling the input resolution from 336 to 1344 \cite{liu2024llavanext}—but at the cost of substantially increasing the number of visual tokens. The resulting token explosion drastically amplifies inference costs, leading to slower throughput, higher latency, and greater GPU memory consumption. Paradoxically, however, only a small fraction of these tokens meaningfully contribute to the final output. For example, in LLaVA-1.5 \cite{llava1.5}, the average attention weight assigned to image tokens in the LLM accounts for only about 0.2$\%$ of that for text tokens. This observation highlights significant redundancy: despite processing vast numbers of visual tokens, LVLMs devote most of their computation to tokens that are largely ignored by the language model.

\begin{figure}[t]
  \centering
  \includegraphics[width=0.5\textwidth]{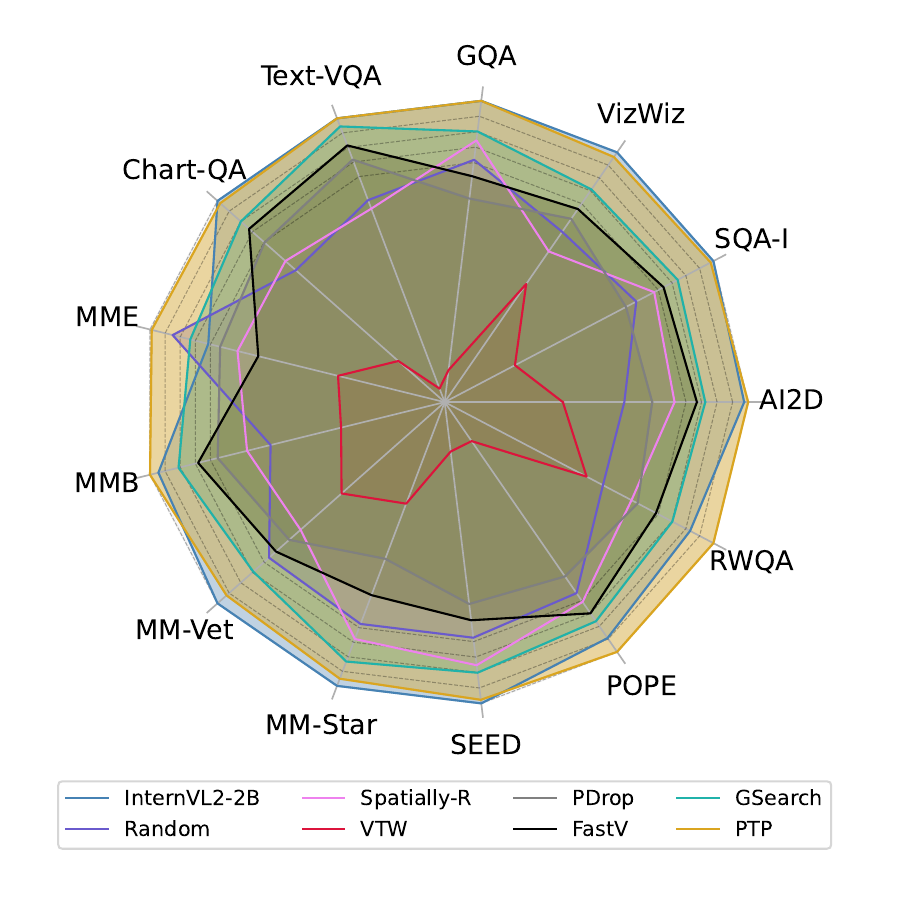}
  \caption{Performance comparison with the baseline model and other token pruning methods across 13 LVLM benchmarks.}
  \label{intro-radar}
\end{figure}

\begin{figure*}[t]
  \centering
  \includegraphics[width=0.7\textwidth]{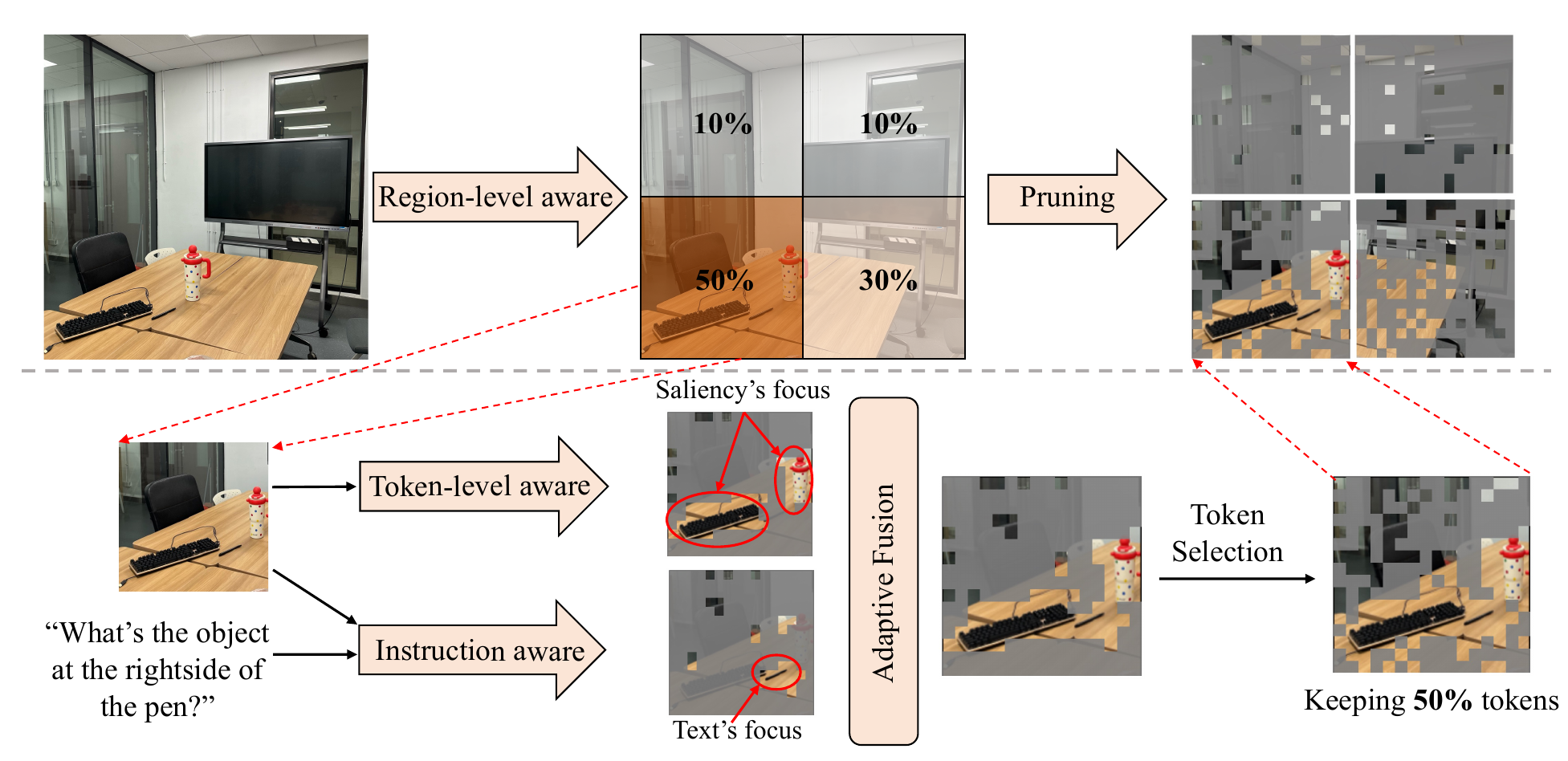}
  \caption{Pipeline of the proposed Pyramid Token Pruning (PTP) method. Following bottom-up and top-down pruning, tokens at critical regions are preserved.}
  \label{intro-pipeline}
\end{figure*}

To address the inefficiency of LVLMs, prior studies have explored visual token compression from multiple perspectives. Projector-side methods, such as TokenPacker \cite{li2024tokenpacker} and HoneyBee \cite{cha2024honeybee}, redesign the visual projector to condense or abstract visual features. While these approaches improve the trade-off between quality and efficiency, they require additional training and model-specific modifications, which limits their plug-and-play applicability across LVLMs. In contrast, training-free methods like PruMerge \cite{Llava-prumerge} and FasterVLM \cite{zhang2024fastervlm} prune visual tokens before they are fed into the LLM, typically relying on CLS attention from the vision encoder. By operating entirely at the pre-LLM stage, these methods can reduce redundancy early, but their text-agnostic design risks discarding task-critical evidence when different instructions focus on different image regions. Another line of training-free work, including FastV \cite{fastv}, SparseVLM \cite{zhang2024sparsevlm}, and VTW \cite{VTW}, prunes tokens inside the LLM based on text-to-vision attention. Although these methods leverage cross-modal information, they overlook the rich visual saliency available in the encoder. More recently, optimization-based approaches such as G-Search \cite{Gsearch} attempt to automatically determine layer-wise token retention rates, but they still depend on validation sets and model-specific heuristics. These limitations motivate our Pyramid Token Pruning (PTP), a plug-and-play, training-free scheme that unifies bottom-up visual saliency (at both region and token levels) with top-down instruction guidance.

The design of PTP is inspired by human visual cognition. As illustrated in Fig.~\ref{intro-pipeline}, PTP adopts a three-stage, coarse-to-fine pruning pipeline consisting of region allocation, token selection, and instruction-aware refinement. Following the observation that human attention first anchors on salient regions and then zooms into finer details \cite{pan2017salgan,liang2025globalsementic, ge2022tcnet}, PTP begins with region-level pruning. In high-resolution LVLMs, cropped sub-images often contribute unevenly; thus, PTP assigns each sub-image a vision-only saliency score to quantify its “visual interest” and allocates token budgets accordingly, giving salient regions larger quotas. At the next stage, token-level pruning is performed within each tile, since even salient regions may contain redundant tokens. This coarse-to-fine procedure prunes uninformative regions more aggressively while preserving the most useful tokens from informative areas. Because bottom-up saliency alone may overlook task-relevant evidence, PTP incorporates a final instruction-aware refinement. After saliency-based pruning, the textual query provides top-down context to identify tokens strongly associated with the instruction, ensuring that critical evidence is not inadvertently discarded. Through this adaptive fusion of bottom-up saliency and top-down guidance, PTP achieves a task-aware pruning process, entirely without retraining or backbone modification. By integrating coarse-to-fine visual cues with instruction relevance, PTP guarantees that the retained tokens are both visually salient and aligned with the current task.

We evaluate PTP on 13 diverse benchmarks using InternVL2-2B and InternVL2-8B \cite{chen2024farinternvl1_5} as the baseline LVLMs. The comparison with existing SOTA methods is shown in Fig.~\ref{intro-radar}. Experiments show that PTP can accelerate inference dramatically with negligible performance loss: we manage to prune a large fraction of visual tokens (e.g. reducing token count by 50$\%$) while keeping accuracy within a few points of the unpruned model on all tasks. These results confirm that PTP effectively preserves the informative content required by LVLMs while addressing the key bottlenecks of high-resolution inference. Detailed ablations further reveal that the balance between saliency-based and instruction-based pruning is task-dependent: bottom-up saliency is sufficient for OCR-style tasks, whereas instruction guidance plays a greater role in open-domain scene understanding such as RealWorldQA \cite{realworldqa}. Beyond demonstrating its effectiveness, these findings provide insights into what kinds of visual tokens are truly necessary for different multimodal tasks.

In summary, our work makes the following contributions:
\begin{itemize}
    \item \textbf{Bottom-up Token Pruning:} We design a pyramid-style pruning mechanism that leverages visual saliency at both region and token levels. This approach effectively removes redundant visual tokens and significantly improves the efficiency of high-resolution LVLM inference, without requiring model modifications or re-training. 
    \item \textbf{Top-down Token Pruning:} We introduce an instruction-aware pruning stage that complements visual saliency with textual context. By incorporating user queries into the pruning process, PTP preserves task-relevant tokens that purely bottom-up methods would otherwise discard, achieving a better trade-off between efficiency and accuracy.
    \item \textbf{Comprehensive evaluation and insights:} We unify the bottom-up and top-down strategies into the PTP framework and demonstrate its effectiveness on InternVL2-2B and InternVL2-8B, achieving substantial speedups with negligible performance loss across 13 diverse benchmarks. Moreover, our ablations reveal task-dependent preferences for saliency-based versus instruction-guided pruning, offering new insights into what visual tokens LVLMs truly require.
\end{itemize}

\section{Related work}

\subsection{Development of Large Vision-Language Models}

Large vision-language models (LVLMs) couple vision encoders with large language models to enable multimodal understanding. Early works such as LLaVA \cite{liu2024llava}, ShareGPT4V \cite{chen2025sharegpt4v}, and InstructBLIP \cite{instructblip} established this paradigm by connecting pre-trained visual encoders (e.g., CLIP-based ViTs \cite{clip,vit,xiao2024binaryvit, song2025mvclip}) to LLMs, thereby supporting open-ended visual question answering and dialogue. Subsequent models improved performance by scaling both model capacity and training data. For example, the InternVL family \cite{chen2024internvl,chen2024farinternvl1_5} aligns a 6B-parameter vision backbone with LLMs, narrowing the gap to proprietary systems such as GPT-4V \cite{gpt4}. More recent variants \cite{chen2023pali,monkey,dong2024internlmxcomposer4khd,2024idefics2,qwenvl} emphasize high-resolution inputs to capture fine-grained details. These methods commonly adopt sub-image partitioning to divide large images into local tiles that can be encoded individually, achieving enhanced fine-grained perception. However, encoding high-resolution inputs into thousands of visual tokens introduces severe efficiency challenges during inference, motivating research into token reduction techniques that preserve performance while lowering computational cost.

\subsection{Token Purning in Large Vision-Language Models}

To alleviate the efficiency bottlenecks of LVLMs, a growing body of work has investigated visual token compression strategies. These methods can be categorized into four directions. 

\textbf{Projector-side approaches.} Token-level reduction can be achieved by modifying the vision-to-language projector. For example, TokenPacker \cite{li2024tokenpacker} introduces a coarse-to-fine packing strategy that aggregates regional cues into compact tokens, while HoneyBee \cite{cha2024honeybee} designs locality-enhanced abstractors based on convolution and deformable attention. Such designs improve the quality–efficiency trade-off by explicitly controlling the number of output tokens, but they require additional training and model-specific components, which limits their portability across different LVLM architectures. 

\textbf{Training-free compression before the LLM.} A second line of work prunes visual tokens at the output of the vision encoder before feeding them into the LLM. PruMerge \cite{Llava-prumerge} identifies important tokens based on sparsity in CLS-to-patch attention and merges the remaining ones into the retained set. Similarly, FasterVLM \cite{zhang2024fastervlm} relies on CLS attention to select the most valuable visual tokens. These methods can effectively reduce redundancy early in the pipeline, but they remain text-agnostic, and thus cannot adapt token selection to the current instruction. As a result, they may inadvertently discard tokens that are critical for certain queries, especially when different instructions focus on different regions of the same image.  

\textbf{Training-free pruning inside the LLM.} Another family of methods targets the LLM layers directly. FastV \cite{fastv} prunes visual tokens with low attention scores after early layers; SparseVLM \cite{zhang2024sparsevlm} recycles decoder cross-attention to adaptively prune instruction-irrelevant tokens across layers; and VTW \cite{VTW} adopts a more aggressive scheme by discarding all vision tokens beyond a selected depth. While these strategies exploit text–vision interactions, they often rely on layer-specific heuristics and overlook the pure visual saliency encoded in the vision backbone. Consequently, they may remove visually important evidence that has not yet been linked to textual cues.  

\textbf{Optimization-based schedules.} More recently, automatic strategies such as G-Search \cite{Gsearch} have been proposed. By analyzing the correlation of attention rankings across layers, G-Search employs greedy search or parametric functions (e.g., P-Sigmoid) combined with Bayesian optimization to determine layer-wise token retention rates. Although these methods are parameter-free in principle, they still require validation sets and model-specific tuning, limiting their generalizability.  

While prior pruning strategies demonstrate the feasibility of reducing token redundancy, they are either training-dependent, text-agnostic or optimization-dependent. These limitations highlight the need for a principled and versatile framework that can unify bottom-up visual saliency.

\section{Method}

\begin{figure*}[t]
\centering
\includegraphics[width=1\textwidth]{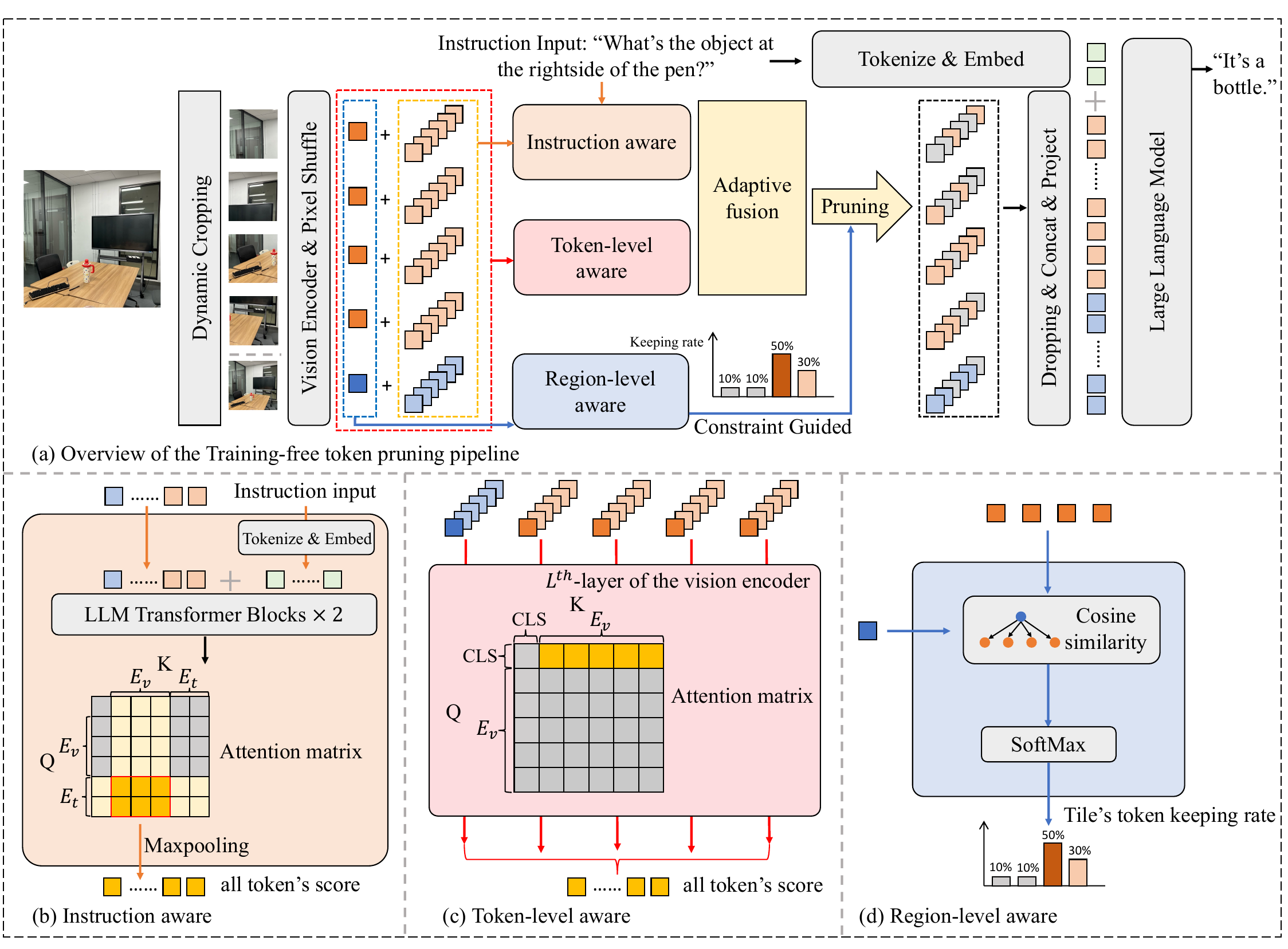}
\caption{(a) Overall workflow of PTP. (b) Top-down instruction-guided importance computation. (c) Bottom-up token-level importance computation. (d) Bottom-up region-level importance computation.}
\label{Overall-frame}
\end{figure*}

We build our method on InternVL2 \cite{chen2024farinternvl1_5}, which follows a standard high-resolution image processing pipeline. A high-resolution image is first partitioned by a dynamic cropping module into multiple sub-images, accompanied by a downsampled thumbnail of the entire image. All image tiles are then resized to a target resolution of 448. The vision encoder processes each tile to produce a set of patch embeddings and a global CLS embedding, which are subsequently mapped through a vision–language projector into the LLM’s embedding space. The LLM consumes these visual tokens together with text tokens to generate responses. 

Our proposed Pyramid Token Pruning (PTP) is inserted as a plug-and-play module between the vision encoder and the LLM, filtering out uninformative visual tokens without any training. As illustrated in Fig.~\ref{Overall-frame}, PTP performs a three-tier importance estimation to identify the most relevant tokens. Tokens are pruned according to these importance scores, enabling substantial acceleration of multimodal inference while preserving model fidelity. Notably, PTP is model-agnostic and requires no additional training or fine-tuning, making it straightforward to integrate into existing LVLM pipelines.

\subsection{Region-Level Importance Scoring}
We begin by evaluating the relative importance of each sub-image at the region level. The input image is first partitioned into an $m \times n$ grid of sub-images, together with a downsampled global thumbnail. Each sub-image is independently encoded by the vision encoder, yielding a sequence of patch embeddings and a [CLS] token that summarizes the region. Let $\mathbf{cls}^{(i)}$ denote the CLS embedding of the $i$-th region, and $\mathbf{cls}^{(g)}$ denote the CLS embedding of the global image. To estimate the semantic contribution of region $i$ to the overall scene, we compute the cosine similarity between its regional CLS embedding and the global CLS embedding:
\begin{equation}
a_i = \frac{\mathbf{cls}^{(i)} \cdot \mathbf{cls}^{(g)}}{\|\mathbf{cls}^{(i)}\| \, \|\mathbf{cls}^{(g)}\|}
\end{equation}
This score serves as the region-level saliency score, reflecting how strongly the $i$-th sub-image aligns with the global semantics of the scene. Next, we normalize the region saliency scores ${a_i}$ with a softmax function to obtain a probability distribution over region importance. Based on a target overall pruning ratio $r$, we then allocate to each sub-image a token budget $R_i$ as:
\begin{equation}
R_i = (1-r) \cdot T \cdot \mathrm{softmax}(a_i),
\end{equation}
where $T$ denotes the total number of visual tokens before pruning. In this formulation, sub-images with higher saliency scores are assigned larger token budgets, ensuring that visually important regions preserve more tokens.

\subsection{Token-Level Bottom-Up Scoring}
After allocating tokens across regions, PTP further evaluates the importance of individual patch tokens within each region using a bottom-up attention-based score. The goal is to identify which patches contribute most to the region’s overall representation. To this end, we leverage the self-attention mechanism of the ViT. Specifically, from the $L$-th transformer layer of the visual encoder, we extract the attention weights from the region’s CLS token to all patch tokens. Formally, for region $i$, the bottom-up score of its $j$-th patch token is defined as the CLS-to-patch attention coefficient at layer $L$:
\begin{equation}
b_j^{(i)} = \mathrm{Attn}_{\mathrm{cls} \rightarrow \mathrm{patch}}^{(L)}(j).
\end{equation}
Here, $\mathrm{Attn}^{(L)}_{\mathrm{cls} \rightarrow \mathrm{patch}}(j)$ denotes the attention weight assigned by the CLS token to patch $j$ at layer $L$. A higher score $b_j^{(i)}$ indicates that patch $j$ has a stronger influence on the regional representation and therefore carries more salient visual information. By computing $b_j^{(i)}$ for all patch tokens in region $i$, we obtain a ranking that reflects their intrinsic visual significance within that region.

\begin{table*}[t]
\centering
\fontsize{7}{9}\selectfont
\caption{\textbf{Comparison of different token pruning methods based on InternVL2-2B and InternVL2-8B across 13 LVLM benchmarks.} "Avg. Acc." denotes the average performance reservation relative to the baseline model. The top-2 best scores are highlighted in bold and underline respectively.}
\label{main-result}
\setlength{\tabcolsep}{0.1pt}  
\renewcommand{\arraystretch}{1.2}
\begin{tabularx}{\textwidth}{
  l|
  *{4}{>{\centering\arraybackslash}X}|
  *{2}{>{\centering\arraybackslash}X}|
  *{5}{>{\centering\arraybackslash}X}|
  *{2}{>{\centering\arraybackslash}X}|
  >{\centering\arraybackslash}X
}
\toprule
\textbf{Model + Method} &
\multicolumn{4}{c|}{\textbf{General VQA}} &
\multicolumn{2}{c|}{\textbf{OCR}} &
\multicolumn{5}{c|}{\textbf{Comprehensive}} &
\multicolumn{2}{c|}{\textbf{Others}} &
\textbf{Avg. acc. (\%)} \\
& \rotatebox{90}{AI2D} & \rotatebox{90}{SQA-I} & \rotatebox{90}{VizWiz} & \rotatebox{90}{GQA}
& \rotatebox{90}{Text-VQA} & \rotatebox{90}{Chart-QA}
& \rotatebox{90}{MME} & \rotatebox{90}{MMB} & \rotatebox{90}{MM-Vet} & \rotatebox{90}{MM-Star} & \rotatebox{90}{SEED}
& \rotatebox{90}{POPE} & \rotatebox{90}{RWQA} &
\\
\midrule
InternVL2-2B\cite{chen2024farinternvl1_5} & 72.5 & 94.1 & 29.4 & 57.3 & 72.0 & 71.4 & 1821.7 & 72.5 & 39.6 & 50.0 & 70.9 & 85.4 & 56.7 & 100.0 \\
+ Random ($r$=0.5) & 69.7 & 90.1 & 26.7 & 55.9 & 57.2 & 51.6 & \underline{1842.2} & 69.8 & 34.2 & 47.8 & 68.5 & 83.9 & 53.1 & 92.3 \\
+ Spatially-R ($r$=0.5) & 70.8 & 90.4 & 26.5 & 56.2 & 57.1 & 52.5 & 1780.1 & 71.3 & 33.2 & 48.0 & 69.7 & 84.5 & 54.7 & 92.6 \\
+ VTW\cite{VTW} & 66.8 & 75.2 & 22.5 & 40.8 & 10.7 & 10.2 & 1596.3 & 64.2 & 22.7 & 37.2 & 52.6 & 64.7 & 52.3 & 69.2 \\
+ PDrop\cite{pdrop} & 70.6 & 89.9 & 26.9 & 55.3 & 70.2 & 64.3 & 1800.1 & 71.5 & 33.9 & 41.6 & 64.8 & 82.1 & 54.8 & 93.8 \\
+ FastV\cite{fastv} ($r$=0.5) & 71.6 & 91.0 & 27.1 & 55.8 & 70.7 & 66.1 & 1774.2 & 71.8 & 34.1 & 46.5 & 68.1 & 84.8 & 55.7 & 95.8 \\
+ GSearch\cite{Gsearch} & \underline{71.9} & \underline{92.5} & \underline{28.3} & \underline{56.4} & \underline{71.6} & \underline{68.2} & 1831.7 & \underline{71.9} & \underline{36.7} & \underline{48.8} & \underline{70.3} & \underline{85.2} & \underline{56.5} & \underline{98.1} \\
+ PTP ($r$=0.5, $\alpha$ =0.5) & \textbf{72.6} & \textbf{93.9} & \textbf{29.2} & \textbf{57.3} & \textbf{72.0} & \textbf{70.7} & \textbf{1879.7} & \textbf{72.8} & \textbf{38.6} & \textbf{49.6} & \textbf{70.7} & \textbf{86.5} & \textbf{57.5} & \textbf{99.8} \\
\midrule
InternVL2-8B\cite{chen2024farinternvl1_5} & 82.4 & 97.0 & 32.9 & 62.7 & 76.6 & 82.6 & 2205.3 & 81.8 & 60.0 & 62.0 & 76.1 & 86.7 & 65.5 & 100.0 \\
+ VTW\cite{VTW} & 81.8 & 62.3 & 23.0 & 56.6 & 18.9 & 16.3 & 2195.1 & \textbf{81.6} & 38.8 & 45.9 & 46.3 & 85.8 & 63.4 & 74.5 \\
+ PDrop\cite{pdrop} & 80.4 & 92.1 & 29.9 & 62.3 & 75.6 & \underline{81.6} & 2193.1 & 81.4 & 56.3 & 60.1 & 71.6 & 86.7 & 64.6 & 97.2 \\
+ FastV\cite{fastv} ($r$=0.5) & 81.1 & 93.2 & 30.3 & 62.0 & 75.6 & 80.2 & 2214.2 & 81.2 & \underline{56.6} & 60.5 & 72.4 & 86.5 & 64.6 & 97.6 \\
+ GSearch\cite{Gsearch} & \underline{82.1} & \underline{94.3} & \underline{31.0} & \underline{62.6} & \underline{76.0} & 81.1 & \underline{2216.7} & 81.4 & 55.9 & \underline{60.8} & \underline{73.2} & \underline{86.9} & \underline{65.1} & \underline{98.0} \\
+ PTP ($r$=0.5, $\alpha$ =0.5) & \textbf{83.8} & \textbf{96.7} & \textbf{33.1} & \textbf{63.1} & \textbf{76.4} & \textbf{82.3} & \textbf{2220.5} & \underline{81.5} & \textbf{56.9} & \textbf{61.4} & \textbf{75.1} & \textbf{87.2} & \textbf{65.3} & \textbf{99.7} \\
\bottomrule
\end{tabularx}
\end{table*}

\subsection{Instruction-Guided Top-Down Scoring}
While the previous scores reflect image-driven saliency, PTP also incorporates top-down guidance from the textual instruction or query. After visual tokens are projected into the LLM’s embedding space, they are concatenated with instruction tokens and jointly processed by the LLM. To assess instruction relevance, we examine the attention weights assigned from instruction tokens to visual tokens in the early layers of the LLM, where cross-modal grounding typically emerges. In particular, we focus on the second transformer block and collect the attention from each instruction token to each visual token. For a given visual token $j$, its instruction-guided importance $c_j$ is defined as the maximum attention it receives from any instruction token:

\begin{equation}
\label{equation4}
c_j = \max_{q \in Q} \mathrm{Attn}_{q \rightarrow j},
\end{equation}

where $Q$ is the set of all instruction token indices. In other words, $c_j$ measures how strongly at least one word in the instruction attends to visual token $j$. If the user’s prompt focuses on a particular object or region, the corresponding visual tokens will achieve high $c_j$ scores. By obtaining $c_j$ for all candidate tokens, PTP captures text-conditioned saliency, ensuring that the visual tokens crucial for answering the query are prioritized.

\subsection{Pyramid Token Pruning}
Finally, we integrate the three importance signals into a unified pruning policy. For token $j$ in sub-image $i$, the scores ${a_i, b_j^{(i)}, c_j}$ are first normalized to the $[0,1]$ range to ensure comparability. The region-level score $a_i$ determines the regional token budget $R_i$, while the token-level saliency $b_j^{(i)}$ and instruction relevance $c_j$ are combined into a single importance score:

\begin{equation}
s_j = \alpha c_j + (1 - \alpha) b_j^{(i)},
\end{equation}

where $\alpha \in [0,1]$ is a hyperparameter controlling the balance between instruction-guided and vision-driven pruning. For example, setting $\alpha = 0$ corresponds to purely bottom-up pruning based on visual saliency, while $\alpha = 1$ relies entirely on instruction guidance. In practice, $\alpha$ is chosen empirically to balance both criteria. For each region $i$, the top-$R_i$ tokens ranked by $s_j$ are retained. For the global thumbnail image, tokens are ranked by their corresponding ${b_j^{(g)}, c_j}$ scores, and the top-$R_g$ tokens are selected, where $R_g$ is the allocated budget for the global view.

In effect, PTP produces a shortened visual token sequence that preserves the critical information required for the current task. By combining coarse region allocation with fine-grained token scoring from both bottom-up and top-down signals, the pyramidal framework efficiently filters visual content, enabling faster inference with negligible impact on model performance.

\section{Experiments}

\begin{table*}[h]
\centering
\fontsize{8}{10}\selectfont
\caption{\textbf{Impact of Vision Encoder Layer Selection.} Analysis of vision encoder layer selection for bottom-up token-level scoring in the PTP method with $r$ and $\alpha$ fixed.}
\label{Layer-results}
\setlength{\tabcolsep}{1pt}  
\renewcommand{\arraystretch}{1.2}
\begin{tabularx}{\textwidth}{
  c|
  *{3}{>{\centering\arraybackslash}X}|
  *{2}{>{\centering\arraybackslash}X}|
  *{3}{>{\centering\arraybackslash}X}|
  *{2}{>{\centering\arraybackslash}X}
}
\toprule
\textbf{Layer} &
\multicolumn{3}{c|}{\textbf{General VQA}} &
\multicolumn{2}{c|}{\textbf{OCR}} &
\multicolumn{3}{c|}{\textbf{Comprehensive}} &
\multicolumn{2}{c}{\textbf{Others}} \\
& SQA-I & VizWiz & GQA
& Text-VQA & Chart-QA
& MME & MMStar & SEED
& POPE & RWQA \\
\midrule
0     & 92.3 & 26.5 & 54.6 & 71.9 & \underline{70.9} & 1852.6 & 47.1 & 68.2 & 84.1 & 55.6 \\
4     & \underline{93.8} & 28.3 & 56.1 & \textbf{72.3} & \textbf{71.0} & 1867.5 & 48.6 & 69.6 & 85.6 & 56.7 \\
8     & \textbf{93.9} & \textbf{29.2} & \textbf{57.3} & \underline{72.0} & 70.7 & \textbf{1879.7} & \textbf{49.6} & 70.7 & \textbf{86.5} & 57.5 \\
12    & 93.7 & \underline{29.0} & \underline{56.9} & 71.2 & 69.3 & 1863.5 & \underline{49.4} & \underline{70.5} & \underline{85.9} & \underline{57.8} \\
16    & 93.6 & 28.6 & 56.5 & 70.7 & 68.6 & 1859.3 & 49.1 & 70.4 & 85.8 & \textbf{57.9} \\
20    & 92.6 & 28.7 & 56.4 & 70.1 & 68.7 & \underline{1872.3} & 49.2 & 70.2 & 85.5 & 56.6 \\
23    & 92.3 & 28.1 & 55.7 & 69.4 & 68.5 & 1865.2 & 48.1 & 70.1 & 85.3 & 56.0 \\
\bottomrule
\end{tabularx}
\end{table*}

\begin{table*}[h]
\centering
\fontsize{8}{10}\selectfont
\caption{\textbf{Analysis of the hyperparameter \bm{$\alpha$}.} The pruning ratio $r=0.5$, and the 11th layer of vision encoder is selected.}
\label{alpha_ablation}
\setlength{\tabcolsep}{1pt}  
\renewcommand{\arraystretch}{1.1}
\begin{tabularx}{\textwidth}{
  c|
  *{3}{>{\centering\arraybackslash}X}|
  *{2}{>{\centering\arraybackslash}X}|
  *{3}{>{\centering\arraybackslash}X}|
  *{2}{>{\centering\arraybackslash}X}
}
\toprule
$\alpha$ &
\multicolumn{3}{c|}{\textbf{General VQA}} &
\multicolumn{2}{c|}{\textbf{OCR}} &
\multicolumn{3}{c|}{\textbf{Comprehensive}} &
\multicolumn{2}{c}{\textbf{Others}} \\
& SQA-I & VizWiz & GQA
& Text-VQA & Chart-QA
& MMB & MME & MMStar
& POPE & RWQA \\
\midrule
0.0 & 92.3 & 27.7 & 55.8 & 69.2 & 66.6 & 71.6 & 1820.1 & 48.1 & 85.0 & 54.4 \\
0.1 & 93.1 & 28.2 & 56.2 & 71.4 & 68.7 & 72.0 & 1858.2 & 48.8 & 86.1 & 55.3 \\
0.2 & 93.3 & 28.4 & 56.5 & \textbf{72.3} & \textbf{70.9} & 72.6 & 1861.5 & 49.1 & \textbf{86.7} & 55.5 \\
0.3 & 93.2 & 28.5 & 56.8 & 71.6 & \underline{70.8} & 72.9 & 1863.9 & 48.9 & 86.3 & 56.0 \\
0.4 & \underline{93.6} & 28.7 & 56.8 & 71.5 & 70.6 & \textbf{73.1} & 1867.6 & \underline{49.5} & \underline{86.6} & 56.2 \\
0.5 & \textbf{93.9} & \textbf{29.2} & \underline{57.3} & \underline{72.0} & 70.7 & \underline{72.8} & \textbf{1879.7} & \textbf{49.6} & 86.5 & 57.5 \\
0.6 & 93.0 & \underline{29.0} & \textbf{57.5} & 71.3 & 69.6 & 72.6 & \underline{1877.4} & 49.0 & 85.9 & 57.7 \\
0.7 & 92.9 & 28.9 & 56.8 & 70.5 & 68.8 & 72.6 & 1874.8 & 48.8 & 85.8 & \underline{57.9} \\
0.8 & 92.7 & 28.5 & 56.9 & 70.0 & 67.2 & 72.4 & 1871.4 & 48.6 & 86.1 & 57.8 \\
0.9 & 92.5 & 28.3 & 56.6 & 69.1 & 66.3 & 72.1 & 1875.7 & 48.5 & 86.1 & \textbf{58.1} \\
1.0 & 92.3 & 28.2 & 56.3 & 68.8 & 65.6 & 72.0 & 1879.3 & 48.6 & 86.0 & 57.6 \\
\bottomrule
\end{tabularx}
\end{table*}

\subsection{Implementation Details}

\paragraph{Model Architecture.}We adopt the InternVL2 series as the baseline high-resolution LVLMs to evaluate the effectiveness of our method. Specifically, InternVL2-2B consists of an InternViT-300M-448px vision encoder, a two-layer MLP as the vision–language projector, and InternLM2-1.8B as the LLM backbone. The larger model InternVL2-8B includes the same InternViT-300M-448px vision encoder, a two-layer MLP projector, and InternLM2.5-7B-Chat as the LLM backbone.

\paragraph{Evaluation.} To comprehensively assess both the effectiveness and efficiency of our proposed token pruning method, we evaluate it on 13 widely used LVLM benchmarks covering diverse task types, grouped into four categories. General VQA includes AI2D~\cite{ai2d}, SQA-I~\cite{sqa}, VizWiz~\cite{vizwiz}, and GQA~\cite{gqa}. OCR-related tasks are represented by TextVQA~\cite{textvqa} and ChartQA~\cite{chartqa}. Comprehensive multimodal benchmarks include MME~\cite{mme}, MMB~\cite{mmb}, MM-Vet~\cite{mmvet}, SEED~\cite{seed}, and MM-Star~\cite{chen2024wemmstar}, which jointly evaluate general multimodal understanding and reasoning capabilities. Finally, Other tasks consist of POPE~\cite{pope}, which measures hallucination in object existence, and RealWorldQA~\cite{realworldqa}, which targets open-domain scene understanding in real-world scenarios. All experiments are conducted on a single NVIDIA L20 GPU with 48GB of VRAM.

\subsection{Main Results}

Table~\ref{Layer-results} reports the performance of PTP when computing bottom-up token-level importance scores from different layers of the vision encoder. Overall, intermediate layers yield the most reliable results. In particular, layer 8 achieves the highest accuracy across the majority of benchmarks, suggesting that it serves as a robust and broadly effective default configuration for PTP.

We also observe task-dependent preferences across layers. OCR-oriented benchmarks such as TextVQA and ChartQA favor earlier layers (e.g., layer 0 or layer 4), likely because these layers capture low-level spatial details that are crucial for text recognition. By contrast, tasks requiring scene-level reasoning, such as RealWorldQA, benefit from deeper layers, which provide more abstract semantic representations. These trends are consistent with prior findings \cite{li2024instruction-guild-multilayer}, which shows that different visual layers contribute unevenly to distinct multimodal task types. Additional analysis is provided in the appendix.

\subsubsection{With InternVL-2B}
As shown in the upper part of Table \ref{main-result}, PTP attains an average normalized accuracy of 99.8\%, nearly identical to the full-token baseline (100.0\%) while dropping 50\% of visual tokens. Remarkably, PTP even surpasses the full model on several benchmarks, such as AI2D (72.6 vs. 72.5), MME (1879.7 vs. 1821.7), MMB (72.8 vs. 72.5), POPE (86.5 vs. 85.4) and RealWorldQA (57.5 vs. 56.7). These gains suggest that pruning away redundant tokens can reduce noise and sharpen the model’s focus on task-relevant content. When compared with other pruning methods, PTP demonstrates clear superiority. It achieves the best accuracy on all 13 benchmarks of different task categories. For example, on VizWiz, methods like VTW and PDrop lose significant accuracy due to over-pruning visually sparse but semantically crucial regions, whereas PTP maintains 29.2, on par with the unpruned model. On POPE, which tests hallucination robustness, PTP improves beyond both the baseline and all competitors, indicating that its joint saliency–instruction design helps preserve evidence relevant to existence queries.

\begin{table*}[h]
\centering
\fontsize{8}{10}\selectfont
\caption{\textbf{Ablation Study.} "Region" denotes the bottom-up region-level scoring; "$b$" refers to the bottom-up token-level scoring; and "$c$" represents the top-down instruction-guided scoring.}
\label{ablation}
\setlength{\tabcolsep}{1pt}
\renewcommand{\arraystretch}{1.5}  
\begin{tabularx}{\textwidth}{
  l|
  *{4}{>{\centering\arraybackslash}X}|  
  *{2}{>{\centering\arraybackslash}X}|  
  *{3}{>{\centering\arraybackslash}X}|  
  *{2}{>{\centering\arraybackslash}X}|  
  >{\centering\arraybackslash}X         
}
\toprule
\textbf{Model} &
\multicolumn{4}{c|}{\textbf{General VQA}} &
\multicolumn{2}{c|}{\textbf{OCR}} &
\multicolumn{3}{c|}{\textbf{Comprehensive}} &
\multicolumn{2}{c|}{\textbf{Others}} &
\multicolumn{1}{c}{} \\
& AI2D & SQA-I & VizWiz & GQA
& Text-VQA & Chart-QA
& MME & MM-Star & SEED
& POPE & RWQA & \multicolumn{1}{c}{\shortstack{\textbf{Avg}\\[-2pt]\textbf{Decline}\\[-1pt](\%)}} \\
\midrule
PTP($r$=0.5, $\alpha$=0.5) & \textbf{72.6} & \textbf{93.9} & \textbf{29.2} & \textbf{57.3} & \textbf{72.0} & \textbf{70.7} & \textbf{1879.7} & \textbf{49.6} & \textbf{70.7} & \textbf{86.5} & \textbf{57.5} & 0.00 \\
w/o Region & \underline{71.9} & \underline{92.1} & \underline{27.9} & \underline{56.4} & 68.5 & \underline{66.0} & \underline{1878.8} & \underline{48.3} & \underline{70.3} & \underline{85.7} & \underline{56.0} & 1.71 \\
w/o Region+$b$ & 71.4 & 91.8 & 27.6 & 55.4 & \underline{68.6} & 65.3 & 1872.4 & 47.6 & 70.0 & 85.2 & 55.6 & 2.45 \\
w/o Region+$c$ & 69.9 & 90.3 & 27.2 & 54.9 & 64.9 & 64.1 & 1823.6 & 47.4 & 68.6 & 83.9 & 55.3 & 4.25 \\
Random & 69.7 & 90.1 & 26.7 & 55.9 & 57.2 & 51.6 & 1842.2 & 47.8 & 68.5 & 83.9 & 53.1 & 7.27 \\
\bottomrule
\end{tabularx}
\end{table*}

\begin{table*}[h]
\centering
\caption{\textbf{Efficiency Analysis of PTP at Different Pruning Ratios on InternVL2-2B.}}
\label{tab:efficiency_analysis}
\setlength{\tabcolsep}{5pt}
\renewcommand{\arraystretch}{1.5}
\begin{tabular}{l|c|c|c|c|c|c|c}
\toprule
\textbf{Method} & \textbf{Tokens}$\downarrow$ & \textbf{Latency Per Token (ms)}$\downarrow$ & \textbf{Total Time (s)}$\downarrow$ & \textbf{TFLOPs}$\downarrow$ & \textbf{GPU Memory (GB)}$\downarrow$ & \textbf{KV-Cache (MB)}$\downarrow$ & \textbf{Performance}$\uparrow$\\
\midrule
InternVL2-2B    & 1792 & 34.2 & 325.7 & 6.40 & 24.6 & 336.0 & 1821.7 \\
PTP-50\%        & 896  & 19.7 & 187.4 & 3.04 & 20.9 & 168.0 & 1879.7 \\
PTP-40\%        & 716  & 18.5 & 176.2 & 2.41 & 20.1 & 134.3 & 1862.1 \\
PTP-30\%        & 537  & 16.2 & 154.0 & 1.79 & 19.2 & 100.7 & 1835.9 \\
PTP-20\%        & 358  & 15.3 & 145.6 & 1.18 & 18.4 & 67.1  & 1821.8 \\
PTP-10\%        & 179  & 14.5 & 138.4 & 0.58 & 17.5 & 33.6  & 1792.4 \\
\bottomrule
\end{tabular}
\end{table*}

\subsubsection{With InternVL-8B}
A similar trend is observed with InternVL2-8B in the bottom part of Table \ref{main-result}. PTP reaches an average normalized accuracy of 99.7\%, matching the full model with only half of the tokens. It  surpasses the full model on AI2D (83.8 vs. 82.4), VizWiz (33.1 vs. 30.9), GQA (63.1 vs. 62.7), MME (2220.5 vs. 2205.3) and POPE (87.2 vs. 86.7). Compared with competitors, PTP achieves best results on 12 of 13 benchmarks. For example, on ChartQA, where fine-grained OCR cues matter, PTP acheives 82.3, much higher than VTW (18.9) or FastV (80.2). On SEED, which requires complex reasoning, PTP also leads with 75.1, showing that its hybrid bottom-up and top-down scoring preserves reasoning-critical tokens better than attention-only heuristics.

Across both model scales, PTP not only preserves near-lossless accuracy under 50\% token pruning but also outperforms the full-token baseline on multiple benchmarks and consistently outperforms competitors. These results confirm that combining region- and token-level saliency with instruction-aware relevance allows PTP to identify the truly informative tokens, leading to both efficiency gains and robustness improvements.

\subsection{Ablation Study}
\subsubsection{Vision Encoder Layer Selection}

\subsubsection{Hyperparameter $\alpha$}

In table \ref{alpha_ablation}, we investigate the effect of the hyperparameter $\alpha$, which controls the trade-off between bottom-up token importance $b$ and top-down instruction-related importance in the PTP framework.

The results show that setting $\alpha=0.5$ yields the best or second-best performance on most benchmarks in the General VQA and Comprehensive categories. This suggests that the tasks which require understanding a broad range of visual scenes and following varied instructions can benefit from a balanced integration of visual saliency and instruction guidance. In other words, a balanced weighting between bottom-up visual features and top-down question cues appears to be effective for general-purpose VQA and comprehensive reasoning tasks. 

In contrast, OCR-focused benchmarks such as TextVQA and ChartQA perform better with smaller $\alpha$ values (e.g., 0.2). These tasks rely heavily on localized and highly salient visual content (e.g., printed text, chart labels), where bottom-up features dominate. In such cases, a lower $\alpha$ better preserves relevant tokens, while top-down attention from the LLM may not effectively highlight these fine-grained visual details. A similar pattern is observed in POPE, where misleading prompts can disrupt top-down attention, making visual saliency the more reliable signal.

On the other hand, the RealWorldQA task shows improved performance at higher $\alpha$ values (0.8). RealWorldQA often involves complex, open-ended queries that are semantically related to the visual elements. In this scenario, the textual instruction could provide more useful guidance than the bottom-up visual saliency. Therefore, emphasizing the top-down signal (high $\alpha$) enables the model to preserve the most important visual tokens, which further explain the superior performance.

Overall, these findings imply that different types of VQA tasks benefit from different balances of bottom-up and top-down influence. Tasks with substantial textual or structured visual content prefer lower $\alpha$ values (relying more on visual saliency), whereas more open-domain or instruction-driven tasks prefer higher $\alpha$ values. This observation suggests that a dynamic fusion strategy, where $\alpha$ is adapted based on task characteristics or input signals, could be a promising direction for future work. Nonetheless, given its consistently strong performance across diverse benchmarks, we recommend $\alpha=0.5$  as a default setting for general-purpose use of the PTP framework.

\begin{figure*}[h]
\centering
\includegraphics[width=0.95\textwidth]{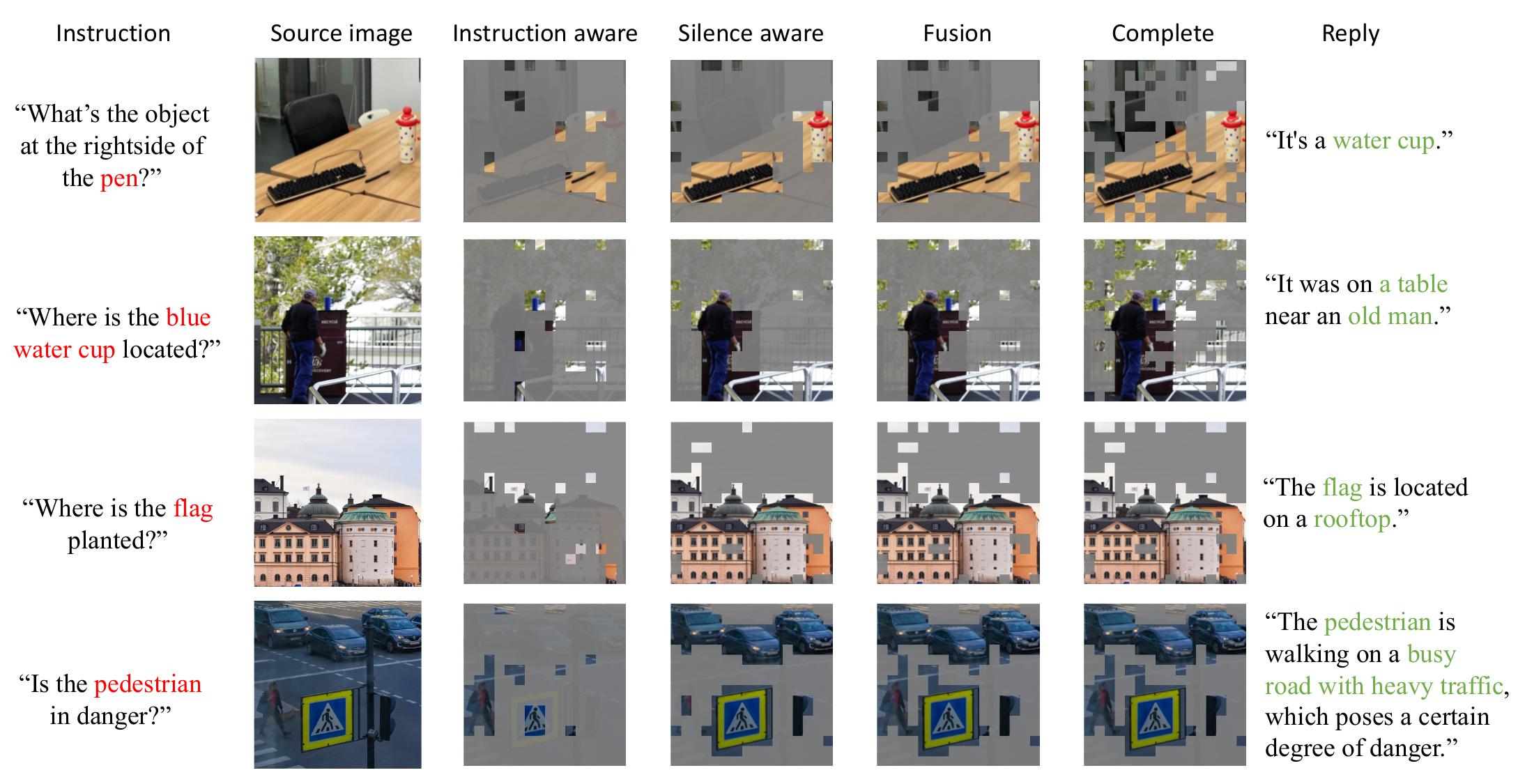}
\caption{Case study of the token pruning results of PTP. The retained tokens (highlighted regions) demonstrate how our method preserves both visually salient content and instruction-relevant information.}
\label{case_study}
\end{figure*}

\subsubsection{Key Component}
Table~\ref{ablation} reports the ablation results isolating the contribution of each component, with the complete PTP framework serving as the baseline.

Removing the region-based token allocation results in a 1.65\% average drop, indicating that accounting for importance differences across local regions is essential. Further excluding the token-level bottom-up score $b$ increases the drop to 2.33\%, confirming that visual saliency cues play a key role in identifying informative tokens. By contrast, removing the instruction-guided score $c$ (together with region allocation) causes a much larger degradation, highlighting the importance of top-down, instruction-driven pruning: leveraging textual guidance ensures that the retained tokens are aligned with task requirements. As a control, random pruning leads to a 7.27\% decrease—far worse than any structured variant.

Overall, these ablation results demonstrate that omitting any component leads to a noticeable performance decline, while the full PTP design effectively combines region-, token-, and instruction-level cues to retain critical visual information. Additional ablation studies are provided in the appendix. 

\subsection{Efficiency Analysis}
Table~\ref{tab:efficiency_analysis} summarizes the efficiency gains of PTP at different pruning ratios on InternVL2-2B. As the pruning ratio increases from 10\% to 50\%, the number of visual tokens decreases from 1792 to 896, yielding substantial improvements in computational efficiency. Specifically, total inference time drops from 325.7 ms to 187.4 ms, and the overall FLOPs reduce from 6.40 TFLOPs to 3.04 TFLOPs, corresponding to a 52.5\% reduction at the 50\% pruning level. Memory consumption also improves: GPU usage decreases from 24.6 GB to 20.9 GB, while the KV-cache size is halved from 336.0 MB to 168.0 MB. These results demonstrate that PTP not only accelerates inference and lowers computational cost but also reduces the memory footprint, thereby improving the deployability of large LVLMs on hardware with more constrained resources.

\subsection{Case Study}
Figure~\ref{case_study} visualizes how PTP prunes and retains tokens under four real queries. Each row shows the source image, the instruction-aware mask (top-down attention from text to vision), the saliency-aware mask (bottom-up visual saliency), their fusion, the complete retained tokens, and the final reply. Across all cases, instruction-only tends to miss necessary visual context, and saliency-only may keep prominent but irrelevant regions. Their adaptive fusion consistently retains just the evidence needed for correct grounding and reasoning, demonstrating that PTP prunes aggressively while preserving instruction-relevant, semantically critical content.

\section{Conclusion}

We have presented PTP, a novel training-free pyramid token pruning framework for high-resolution LVLMs. PTP hierarchically integrates region-level saliency, bottom-up token importance, and top-down instruction guidance to effectively filter visual tokens. This simple yet powerful design delivers substantial efficiency improvements: on InternVL2-2B, PTP reduces FLOPs by 32.2\% while retaining 99.8\% of the original accuracy, and consistently outperforms or matches prior pruning methods across 13 diverse benchmarks. Comprehensive ablation studies confirm that all three components are indispensable for preserving model performance.

Looking ahead, we plan to extend PTP by developing fully training-free, adaptive $\alpha$-fusion strategies that can dynamically balance bottom-up and top-down cues, further enhancing efficiency and robustness for diverse multimodal tasks.

\bibliographystyle{unsrt}
\bibliography{references}

\appendix

\begin{figure*}[h]
  \centering
  \subfigure[Original Image]{
    \includegraphics[width=0.285\textwidth]{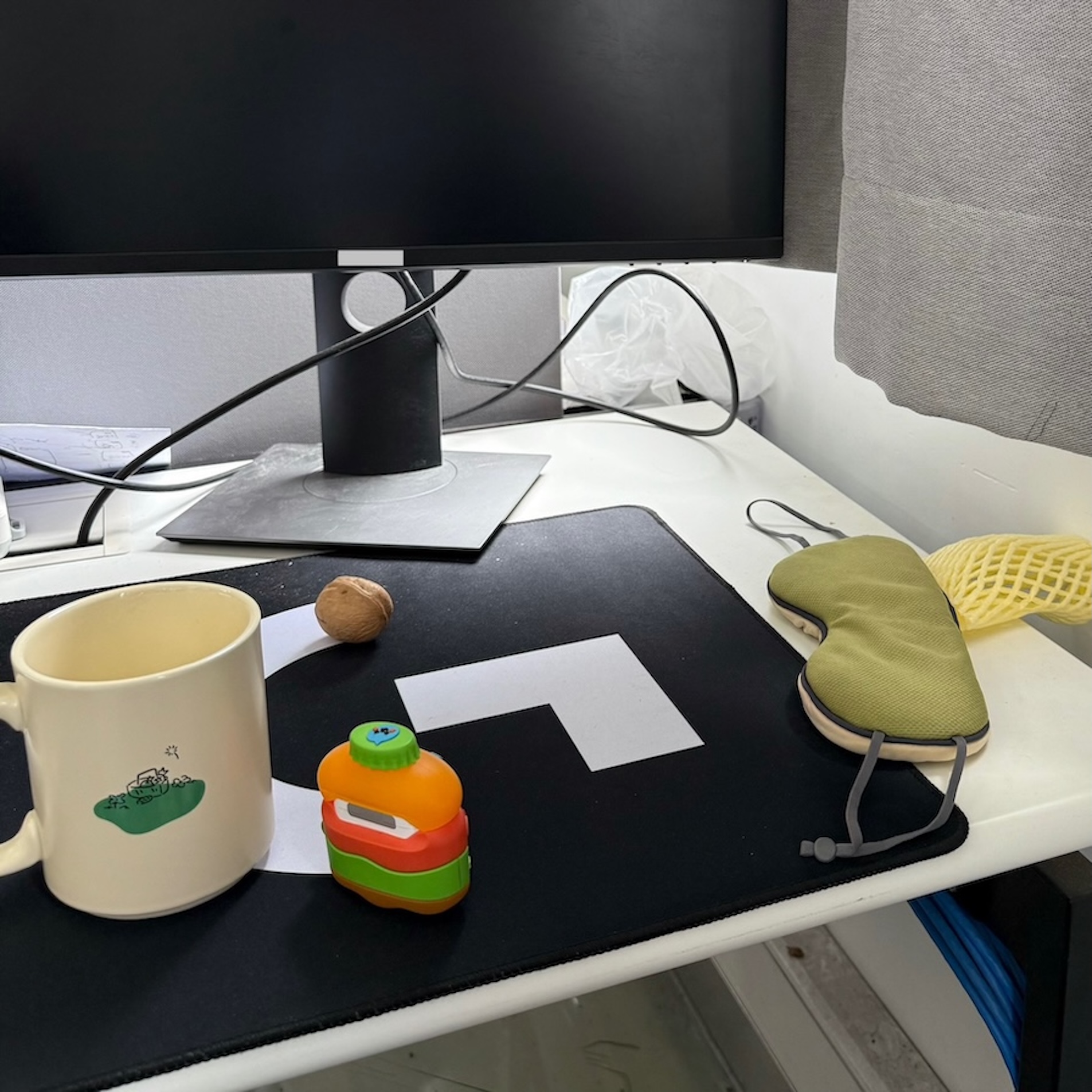}
  }\subfigure[Avg. Attention Head]{ 
    \includegraphics[width=0.3\textwidth]{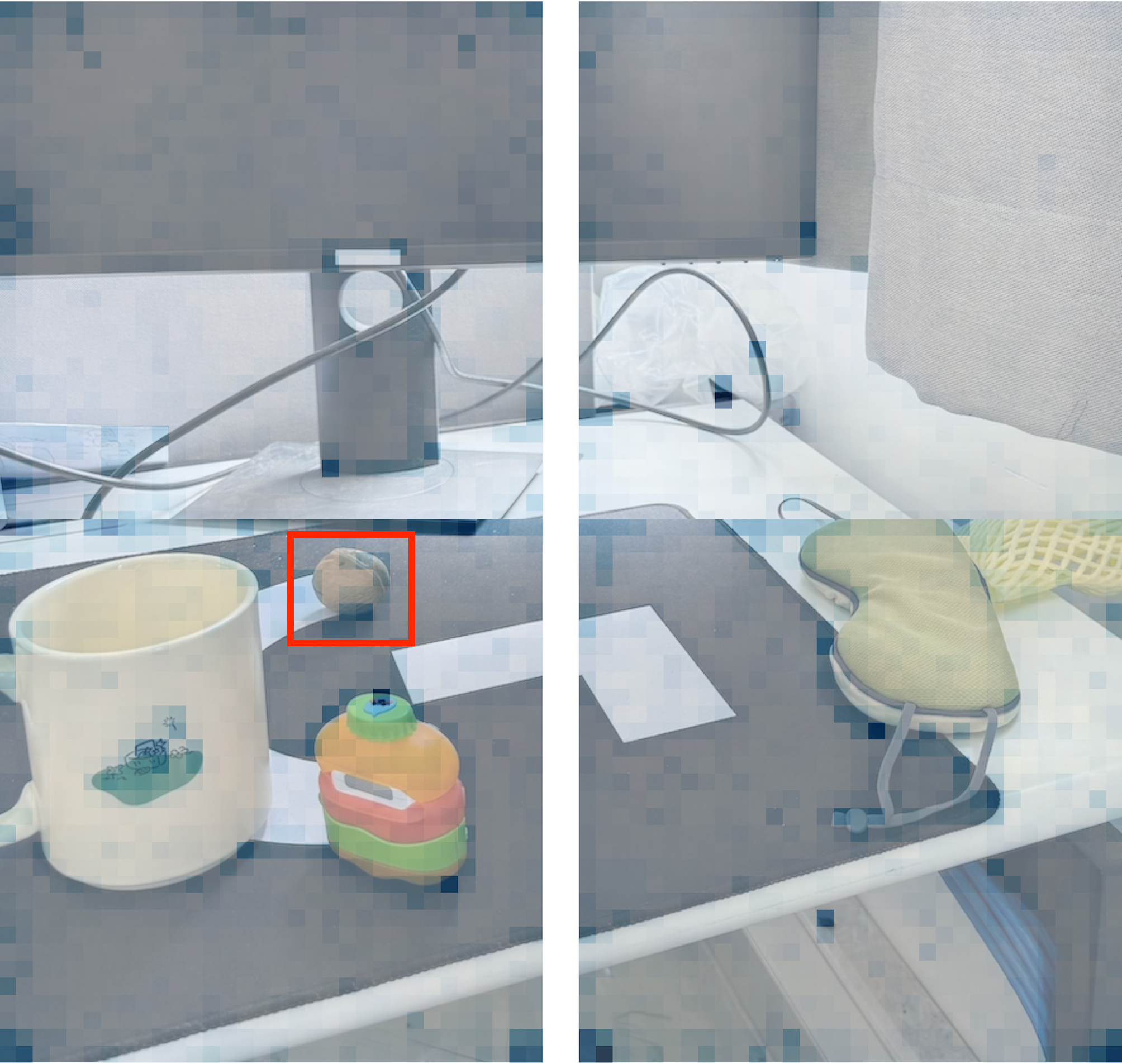}
  }\subfigure[Max. Attention Head]{ 
    \includegraphics[width=0.3\textwidth]{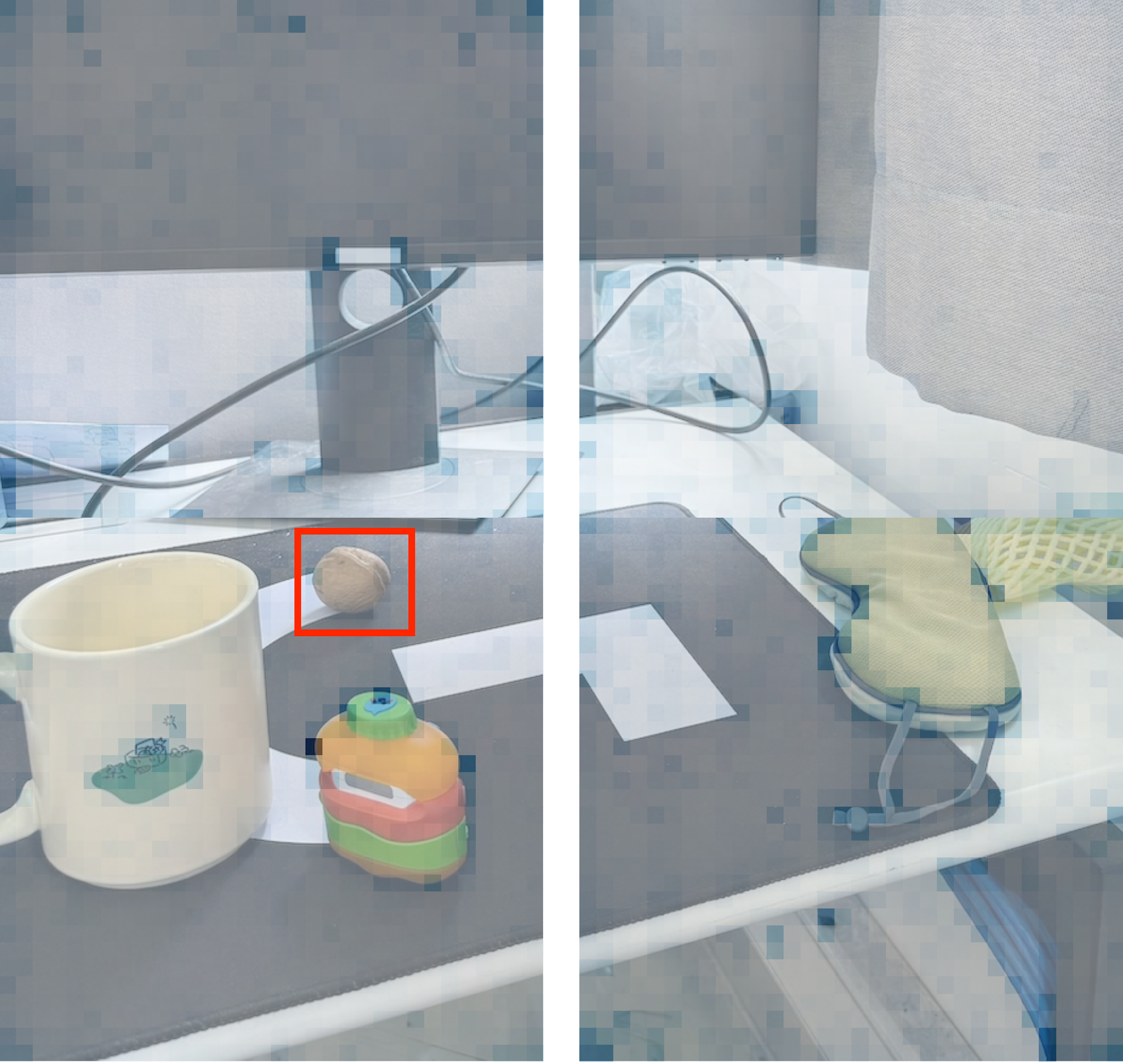}
  }
  
  \caption{Visualization of different attention head selection strategies in the vision encoder.}
  \label{head_analysis}
\end{figure*}

\section*{Analysis of Layer-wise Preferences in the Vision Encoder}
\label{appendixA}

As part of our investigation into bottom-up token-level visual saliency scoring, we further visualize the attention distribution across different layers of the vision encoder, as shown in the Figure \ref{layer-analysis}.
Suppose the high-resolution image $I$ is partitioned into $S$ sub-images $\{I^{(s)}\}_{s=1}^S$. Each sub-image $I^{(s)}$ is encoded into tokens $\mathbf{X}^{(s)} = [\mathrm{CLS}, x_1, \ldots, x_N]$, and is processed by a vision encoder with $L$ layers.

For the $l$-th layer and the $s$-th sub-image, the self-attention matrix is:
\begin{equation*}
    \mathbf{A}^{(l, s)} \in \mathbb{R}^{(N+1) \times (N+1)}
\end{equation*}
where $A_{i, j}^{(l, s)}$ denotes the attention score from token $i$ to token $j$.

We extract the attention scores from the CLS token to all other tokens as:
\begin{equation*}
    \mathbf{a}_{\mathrm{CLS}}^{(l, s)} = \big[ A_{0, 1}^{(l, s)},\, A_{0, 2}^{(l, s)},\, \ldots,\, A_{0, N}^{(l, s)} \big]
\end{equation*}

By visualizing the attention scores $\mathbf{a}_{\mathrm{CLS}}^{(l, s)}$ across different layers of the vision encoder, we observe several distinct patterns:

\begin{itemize}
    \item \textbf{Shallow Layers ($l$ is small):} The attention from the CLS token is primarily distributed along the edges of objects, with the focus being relatively scattered and less structured. This suggests that lower layers are sensitive to local texture and boundary information, but lack object-level semantic understanding.
    
    \item \textbf{Middle Layers:} As we move to the middle layers, the attention of the CLS token becomes significantly more concentrated on the main objects in the scene, while still capturing some key background information. This indicates that the vision encoder at these layers is capable of integrating local features into coherent object-level representations, along with essential contextual cues.
    
    \item \textbf{Deep Layers ($l$ is large):} In deeper layers, we observe an attention drift phenomenon, where the CLS token's focus occasionally shifts away from the main objects, and more attention is paid to a small number of edge or background tokens. This may reflect the model’s attempt to capture global context or may be due to over-smoothing and the loss of discriminative power for object-specific features at higher layers.
\end{itemize}

Combining the above visualization analysis with the empirical results presented in Table \ref{Layer-results}, we select the 11th layer as the default for our PTP method. This layer achieves a balance between focusing on salient object information and capturing key background context, as evidenced by both qualitative visualization and quantitative evaluation. Therefore, the 11th layer is particularly suitable for general-purpose token importance estimation in our framework.

\section*{Visualization of Attention Head Selection Strategy}
\label{appendixB}

When extracting the attention maps from a specific layer of the vision encoder, each token block receives attention scores from multiple attention heads. In our visualization of a sample image, we present both the results of averaging the attention scores across all heads, as well as selecting the maximum score among them. As shown in Figures \ref{head_analysis} (b) and (c), the object "walnut" located at the bottom left of the image receives notable attention under the mean setting, but is almost completely ignored under the max setting. This observation suggests that the mean operation enables a more balanced aggregation of multi-granular information across the image, while the max operation is more likely to overlook certain salient features due to its tendency to focus on the most prominent responses from individual heads.

\section*{Supplementary Ablation Study}
\label{appendixC}

\begin{table}[htbp]
\centering
\small
\caption{Comparison of different instruction-guided token scoring methods on POPE.}
\label{tab:POPE_subtasks}
\setlength{\tabcolsep}{4pt}
\renewcommand{\arraystretch}{1.2}
\begin{tabular}{l|cc}
\toprule
\textbf{Metric} & \textbf{max} & \textbf{mean} \\
\midrule
Overall   & 86.5 & 84.7 \\
Accuracy  & 87.6 & 86.4 \\
Precision & 94.9 & 95.7 \\
Recall    & 79.3 & 75.9 \\
\bottomrule
\end{tabular}
\end{table}

In our proposed PTP method, the instruction-guided importance score $c_j$ is computed as described in Equation \ref{equation4}. Depending on the length and content of the instruction, we select the "max" operation over the instruction-attended attention scores to obtain $c_j$. To evaluate this design choice, we compare the results of using the "max" and "mean" operations for instruction-guided importance calculation, as shown in Table~\ref{tab:POPE_subtasks} and Table~\ref{tab:MME_subtasks}. When adopting the mean operation, $c_j$ is defined as the average attention score from all instruction tokens in $Q$ to token $j$.

\begin{equation*}
c_j = \frac{1}{|Q|} \sum_{q \in Q} \mathrm{Attn}_{q \rightarrow j},
\end{equation*}

Across both the POPE and MME benchmarks, the "max" operation consistently outperforms the "mean" operation. Specifically, for the POPE metric, using "max" yields higher scores in overall performance, accuracy, and recall, while maintaining comparable precision. Similarly, in the MME subtasks, the "max" operation achieves better or comparable results across most subtasks, notably in counting and position. 

\begin{table}[h]
\centering
\small
\caption{Comparison of different instruction-guided token scoring methods on MME.}
\label{tab:MME_subtasks}
\setlength{\tabcolsep}{3pt}  
\renewcommand{\arraystretch}{1.2}
\begin{tabular}{l|cc}
\toprule
\textbf{Task category} & \textbf{max} & \textbf{mean} \\
\midrule
Perception & 1465.56 & 1429.63 \\
Reasoning & 413.71 & 395.00 \\
OCR & 102.50 & 102.50 \\
Artwork & 149.75 & 148.25 \\
Celebrity & 117.75 & 121.18 \\
Code Reasoning & 95.00 & 95.00 \\
Color & 168.33 & 158.33 \\
Commonsense Reasoning & 108.71 & 105.00 \\
Count & 141.33 & 128.33 \\
Existence & 200.00 & 195.00 \\
Landmark & 155.50 & 154.20 \\
Numerical Calculation & 40.00 & 40.00 \\
Position & 151.67 & 141.00 \\
Posters & 128.23 & 129.59 \\
Scene & 150.50 & 151.25 \\
Text Translation & 170.00 & 155.00 \\
\bottomrule
\end{tabular}
\end{table}

These results indicate that the "max" operation is more effective in focusing on instruction-relevant object information, whereas the ``mean'' operation is more likely to be diluted by irrelevant instruction tokens, leading to suboptimal attention allocation. Therefore, adopting the "max" strategy for calculating instruction-guided importance enhances the model's ability to attend to the most critical visual features specified by the instruction.

\section*{Token Budgets Analysis}

\begin{figure}[h]
\centering
\includegraphics[width=0.5\textwidth]{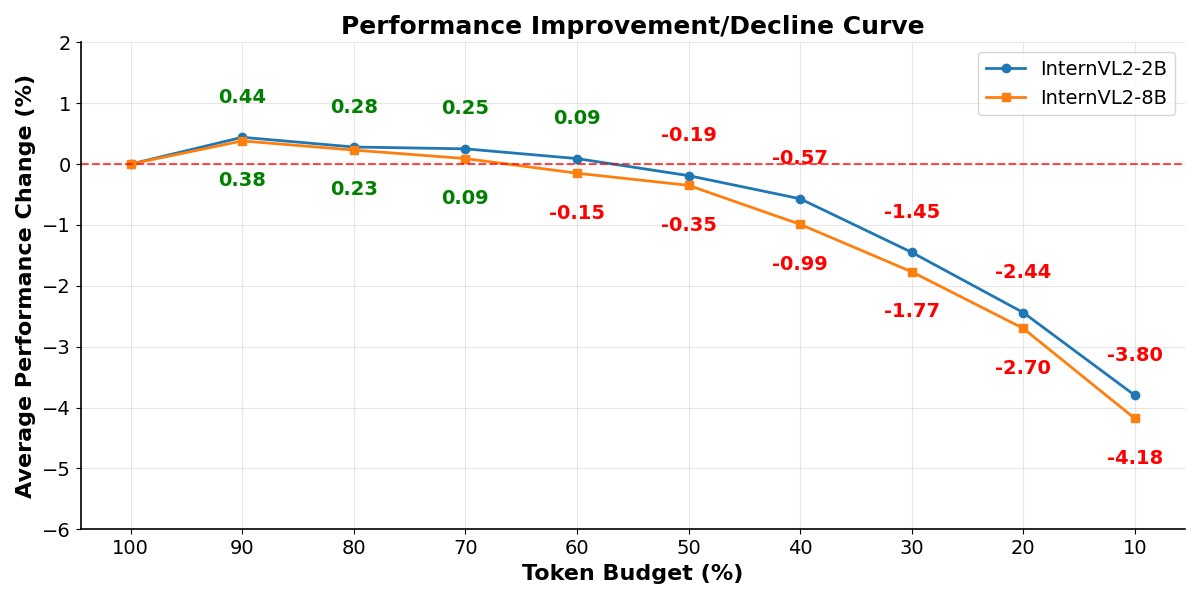}
\caption{Token bugets analysis showing the relationship between pruning ratio and model performance across different benchmarks.}
\label{bugets_analysis}
\end{figure}

As shown in Figure~\ref{bugets_analysis}, we conduct unified token-budget experiments on InternVL2-2B and 8B across five benchmarks: MMStar, VizWiz, SQA, TextVQA, and SEED. In the relatively generous budget regime (70–90\%), our method yields consistent overall gains even after removing a portion of tokens. As compression intensifies, performance declines smoothly; even under extreme compression (10–20\% budget), the average performance drop stays within 5\%, achieving a favorable efficiency–accuracy trade-off. Notably, the 8B curve sits below the 2B curve overall. Because the 8B model has greater capacity, it leverages tokens that appear unimportant to shallower models, mining long-tail signals and cross-token interactions. Removing the same fraction of tokens thus strips more useful information from 8B than from 2B, yielding a relatively larger loss.

\begin{figure*}[t]
\centering
\includegraphics[width=0.65\textwidth]{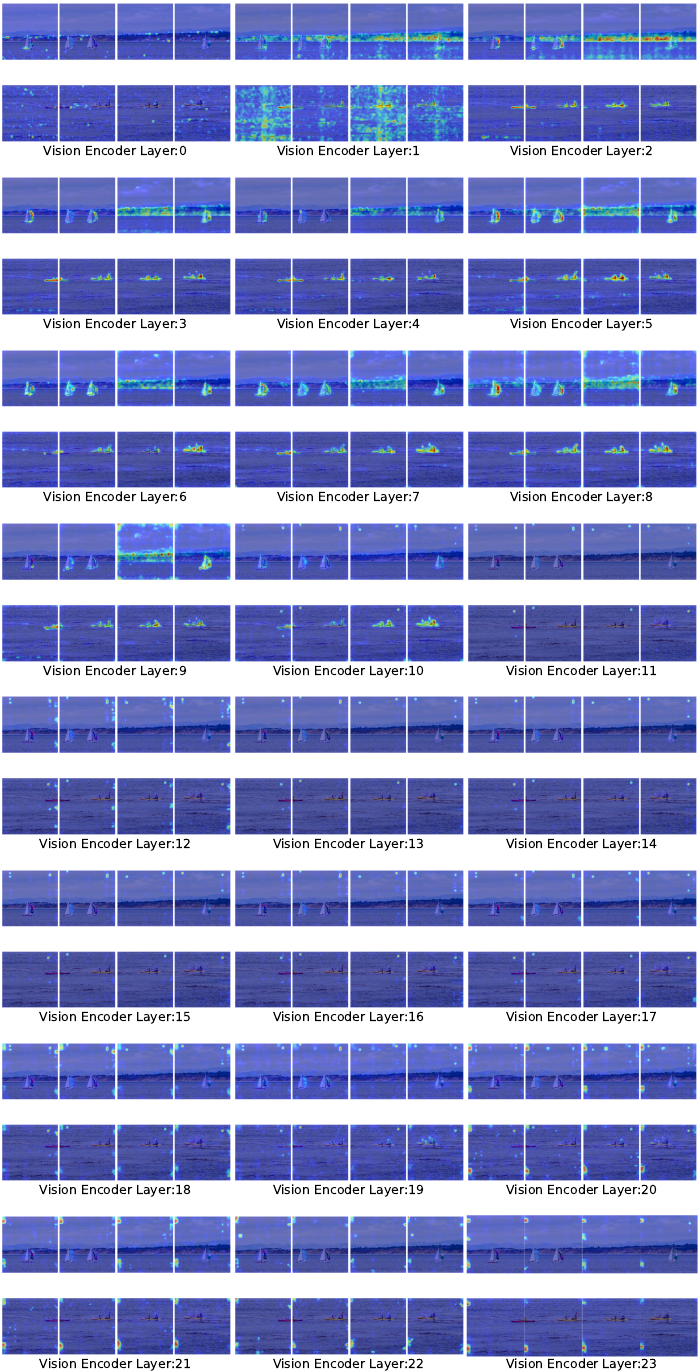}
\caption{Attention analysis across the 24 layers of the vision encoder in InternVL2-2B.}
\label{layer-analysis1}
\end{figure*}

\begin{figure*}[t]
\centering
\includegraphics[width=0.7\textwidth]{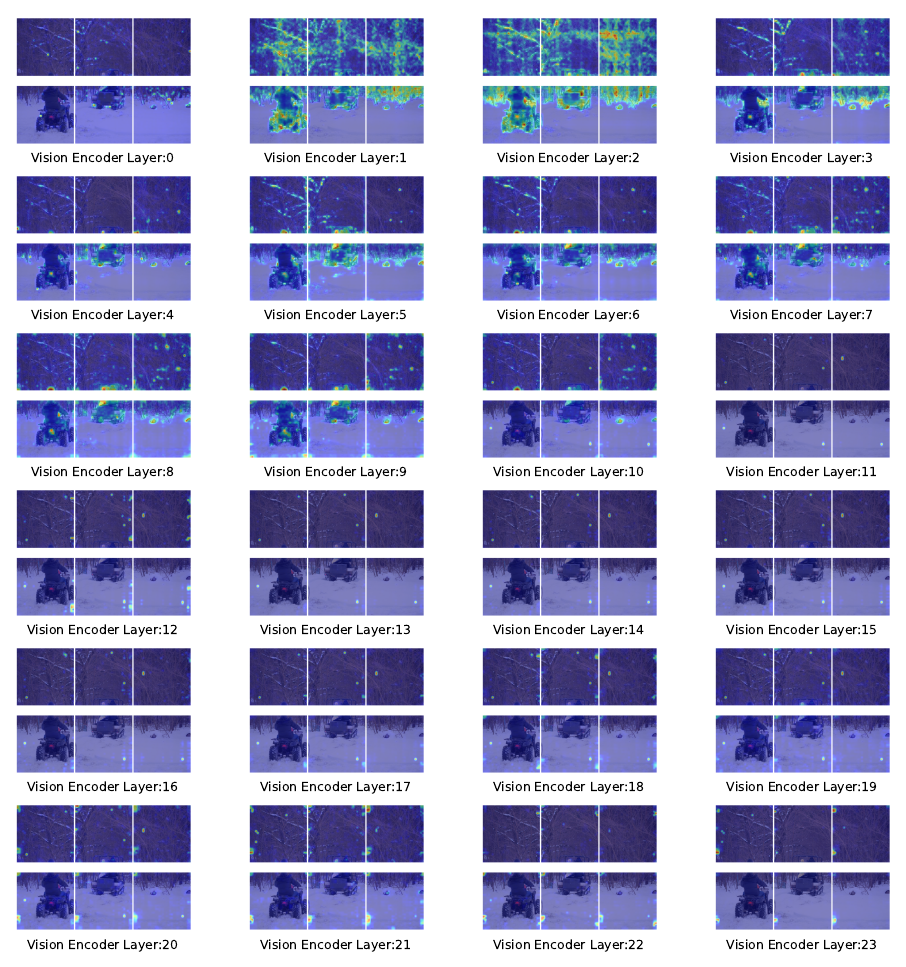}
\caption{Sub-image attention analysis across the 24 layers of the vision encoder in InternVL2-2B.}
\label{layer-analysis2}
\end{figure*}

\begin{figure*}[t]
\centering
\includegraphics[width=0.7\textwidth]{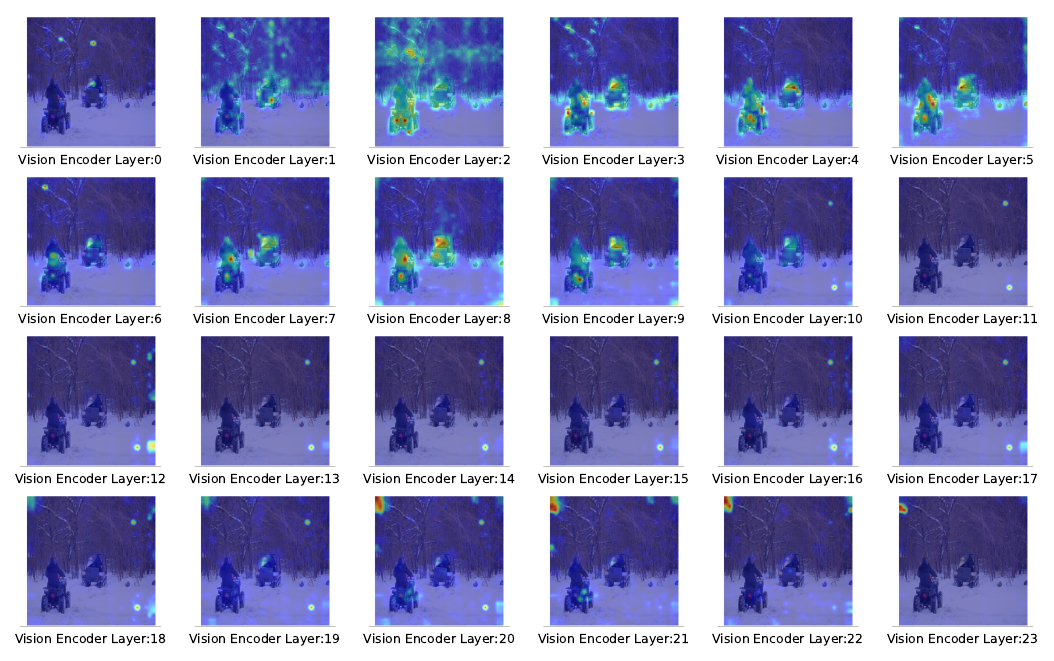}
\caption{Global thumbnail attention analysis across the 24 layers of the vision encoder in InternVL2-2B.}
\label{layer-analysis3}
\end{figure*}

\end{document}